\documentclass[letterpaper]{article} 
\usepackage{aaai2026}  
\usepackage{times}  
\usepackage{helvet}  
\usepackage{courier}  
\usepackage[hyphens]{url}  
\usepackage{graphicx} 
\usepackage{natbib}  
\usepackage{caption} 
\usepackage{algorithm}
\usepackage{algorithmic}

\usepackage{xcolor}
\newcommand{\answerYes}[1]{\textcolor{blue}{#1}} 
 
\newcommand{\answerNA}[1]{\textcolor{gray}{#1}} 
\newcommand{\answerTODO}[1]{\textcolor{red}{#1}}

\usepackage{newfloat}
\usepackage{listings}
\usepackage{graphicx,multirow,xspace,longtable}
\usepackage{tabularx,booktabs,blindtext,url}
\usepackage{makecell,amsmath,rotating, amsfonts}
\usepackage[T1]{fontenc}
\usepackage[utf8]{inputenc}
\usepackage[draft,inline,nomargin,index]{fixme}
\usepackage{marginnote}
\DeclareCaptionStyle{ruled}{labelfont=normalfont,labelsep=colon,strut=off} 
\floatstyle{ruled}
\newfloat{listing}{tb}{lst}{}
\floatname{listing}{Listing}
\usepackage[capitalize,noabbrev]{cleveref}
\usepackage{subcaption}
\newcommand\clearrow{\global\let\rowmac\relax}

\crefformat{section}{\S#2#1#3}
\crefformat{subsection}{\S#2#1#3}
\crefformat{subsubsection}{\S#2#1#3}

\newcommand{\para}[1]{{\vspace{.05in} \bf \noindent #1 }}

\newcommand{\cf}{cf.\ }
\newcommand{\eg}{e.g.,\ }

\newcommand{\ie}{i.e.,\ }

\title{Measuring the Semantic Structure and Evolution of Conspiracy Theories}
\author {
    Manisha Keim,
    Sarmad Chandio,
    Osama Khalid,
    Rishab Nithyanand
}
\affiliations {
    University of Iowa, Iowa City\\
    \{manisha-keim, sarmad-chandio, osama-khalid, rishab-nithyanand\}@uiowa.edu
}

\begin{document}

\maketitle

\begin{abstract}

Research on conspiracy theories largely focusses on belief formation, exposure, and diffusion, while paying less attention to how their meanings change over time.
This gap persists partly because conspiracy-related terms are often treated as stable lexical markers, making it difficult to separate genuine semantic changes from surface-level vocabulary changes.

In this paper, we measure the semantic structure and evolution of conspiracy theories in online political discourse.
Using 169.9M comments from Reddit's r/politics subreddit spanning 2012-2022, we first demonstrate that conspiracy-related language forms coherent and semantically distinguishable regions of language space, allowing conspiracy theories to be treated as \textit{semantic objects}.
We then track how these objects evolved over time using aligned word embeddings, enabling comparisons of \textit{semantic neighborhoods} across time.
Our analysis reveals that conspiracy theories evolve non-uniformly, exhibiting patterns of semantic stability, expansion, contraction, and replacement that are not captured by keyword-based approaches alone.

\end{abstract}
\section{Introduction} \label{sec:introduction}

In November 2016, a gunman enters Comet Ping Pong pizzeria in Washington, D.C USA, firing an AR-15 rifle to ``self-investigate'' claims that the restaurant harbors a child trafficking ring run by Democratic Party officials. The conspiracy theory that motivates this attack, ``Pizzagate'', begins as specific accusations about a single pizza restaurant connected to Hillary Clinton's campaign chairman. By 2020, however, ``Pizzagate'' becomes absorbed into the QAnon conspiracy theory, linking not just to child trafficking but to claims about a global cabal of elites, deep state actors, and institutional control. While the term ``Pizzagate'' remains constant across these years, its meaning, the concepts, actors, and narratives it encompasses, fundamentally transforms and fragments.
This illustrates a fundamental challenge: disambiguating semantic evolution (changes in what a conspiracy theory means) from lexical churn (changes in the vocabulary used to express it). While existing research predominantly focuses on who believes conspiracy theories, how they spread, and what psychological factors drive belief formation \cite{douglas2017psychology, del2016spreading, Samory_Mitra_2018}, understanding how conspiracy theories change in meaning over time remains critically understudied.
Traditional keyword-based approaches \cite{gulordava2011distributional} track specific terms like ``pizzagate'' or ``deep state'', but cannot distinguish between three fundamentally different processes: \textit{(i)} the term itself persists while its underlying meaning transforms, \textit{(ii)} the meaning remains stable while the vocabulary changes entirely, and \textit{(iii)} a single theory fragments into multiple interpretations referenced by the same term. 
By drawing on distributional semantics, we move beyond isolated keywords to use {\em semantic neighborhoods}, the collections of terms that consistently co-occur with conspiracy theories and define what they actually mean in discourse. This approach allows us to define {\em semantic objects}: coherent regions bounded by \textit{semantic neighborhoods}. We use these as stable units of analysis that represent a conspiracy theory's meaning at a given point in space and time. This allows us to track how these objects evolve, whether they remain stable, relocate in semantic space, expand or contract, or fragment into multiple meanings --- while separately measuring vocabulary turnover, revealing evolutionary patterns that prior methods fundamentally cannot capture.
We develop and apply \textit{a semantic object framework} to study conspiracy theory evolution in online political discourse. Using 169.9M comments from Reddit's {\tt r/politics} spanning 2012-2022, we address two research questions:

\begin{itemize}
    \item {\bf RQ1. Is language associated with conspiratorial discourse semantically distinguishable from language used in non-conspiratorial discourse? (\Cref{sec:rq1})}
    Specifically, we ask whether conspiracy-related discourse occupies a coherent and distinguishable region of linguistic expression that is meaningfully separable from non-conspiratorial discourse.
    We focus on establishing this property because semantic evolution can only be meaningfully studied if conspiratorial language forms a coherent and distinguishable semantic region within the discourse under study; without such structure, neither individual conspiracy theories nor their interactions over time can be well defined as they would be diffused over the entire language.
    We address this question by constructing {\em semantic neighborhoods} anchored on 19 distinct conspiracy theories and evaluating their coherence and boundaries using embedding-based cluster analysis and human expert annotations.
    This step establishes whether the \textit{semantic neighborhoods} associated with conspiracy theories can be treated as {\em semantic objects}.

    \item {\bf RQ2. How do the semantics and lexica associated with conspiracy theories evolve over time? (\Cref{sec:rq2})}
    Based on the findings of RQ1, we address this question by building on prior research on diachronic word embeddings \cite{hamilton2016diachronic, kulkarni2015statistically} to create an analysis framework that allows comparisons of \textit{semantic objects}, each object representing the \textit{semantic neighborhood} of an individual conspiracy theory, across time periods.
    These comparisons allows us to reason directly about semantic changes over time without relying on vocabulary shifts alone, thus enabling accurate tracking and characterization of the evolutionary patterns associated with individual conspiracy theories.
    Alongside semantic shifts, we also track changes in the lexica to capture how the language used around these theories changes even when the underlying meaning remains stable.
\end{itemize}

Broadly, our findings shows that conspiracy-related discourse forms coherent and semantically distinguishable clusters, validating the use of conspiracy-related semantic objects as units in our analysis framework.
We also find that conspiracy theories evolves non-uniformly, with some retaining their semantic core over long periods, others either narrowing their semantics to reflect more specific theories or broadening their semantics by absorbing concepts from other theories, and a few undergoing drastic shifts which retain little of their past meanings.
We find that many of these patterns are largely invisible to approaches that are focused on lexicon or keywords alone. We also observe that political scandal conspiracies underwent semantic replacement, whereas elite control conspiracies developed multiple, diverging narratives.

\section{Data and Preprocessing} \label{sec:dataset}
In this section, we describe our dataset, methods for the construction of temporal embeddings, and approach for identifying conspiracy theories and their themes. 

\para{Identifying conspiracy theories and thematic concept labels.} We focus on a fixed set of 19 conspiracy theories that were prominent during the period between 2012-2017 (\cf \Cref{tab:rq1:labels})\footnote{While this is not an exhaustive set, it serves as a proof of concept. The ideas presented here can be extended to any arbitrary conspiracy}. This choice allows us to focus on examining how these theories evolve throughout the three time periods in our study, as social and political dynamics change. 
These conspiracy theories were curated based on prior academic literature \cite{mahl2021nasa, hanley2023golden, samory2018government, Samory_Mitra_2018, 10.1371/journal.pone.0134641, schabes2020birtherism} and media reporting \cite{thomas_2025, uscinski_2016}. These theories include well-established U.S.-centric conspiracy theories (\eg illuminati and chemtrails), event-driven conspiracies (\eg related to the Sandy Hook shooting and Boston bombing), and squarely political conspiracies (\eg Emailgate and Russiagate).
For each of these, we defined a {\em concept label}: a term that encapsulates the core theme of the theory and anchors it in discourse. These labels reflect how the specific theory is commonly referenced in online discussions. For example, the concept label `deep state' refers to conspiracies alleging that a hidden network of government actors secretly controls U.S. policy.
Each concept label serves as an entry point for identifying the surrounding discourse related to a conspiracy theory. This was inspired by prior work \cite{Samory_Mitra_2018, samory2018government} which described similar ``overarching themes'' that structured conspiratorial discourse in online communities. 
We treat these concept labels as anchors rather than exhaustive representations --- \ie we do not assume that a conspiracy theory is fully captured by these labels. Instead, these labels are used to identify and analyze the broader semantic neighborhoods in which conspiratorial discourse appears.

\para{Dataset.} Our baseline dataset consists of 169.9M comments from the {\tt r/politics} subreddit during the period between 2012 and 2022 --- a period that includ multiple highly contentious elections, economic instability, and a global pandemic. 
We select the {\tt r/politics} subreddit for two main reasons. Firstly, it is the largest U.S. political discussion space on Reddit, with continuous engagement over the study period. Secondly, U.S.-related conspiracy theories frequently appear in comments on the subreddit, both directly or through rebuttal. Together, this makes it a suitable venue for studying discourse around conspiracy theories in the mainstream U.S. political context.
We intentionally avoid conspiracy-focused communities (\eg {\tt r/conspiracy}) because they represent niche communities where conspiracy theories are normalized and rarely critiqued or contested. Our goal is to understand the evolution of conspiracy theories in mainstream discourse while accounting for the articulation, rebuttal, and reframing of conspiracy narratives over time. 
%

\para{Constructing temporal word embeddings.} Prior to constructing embeddings, we preprocess each comment by removing URLs, eliminating stopwords, standardizing casing, and performing lemmatization. We construct word embeddings to capture the semantic relationships between words, enabling us to model how the meanings of terms and entire conspiracy theories evolve over time. We also partition comments into three U.S.-specific political time periods: 2012-2014 (a period of fringe conspiratorial discourse marked by events such as Sandy Hook and ``crisis actor''), 2015-2019 (mainstreamed and politically salient conspiracies related to the 2016 elections), and 2020-2022 (pandemic-driven conspiratorial narratives), each reflecting contexts\footnote{We acknowledge that the time spans are not equal because our focus is on distinct political periods in U.S. politics. However, our method generalizes to any time period.}.

Next, to identify multi-word expressions, we use a conditional probability-based approach, motivated by the observation that many conspiratorial terms derive their meaning from short word combinations rather than individual words. Also, we focus on bigrams because they are expressive enough to capture these meanings (e.g., false flag, crisis actor), while remaining frequent enough to allow statistical estimation. For each bigram ($w_1$, $w_2$), we compute the conditional probability $Pr(w_2|w_1)$. We calculate a $z$-score for all bigrams, relative to this distribution of conditional probabilities, and keep those whose $z$-score exceed 1.96 (approximately the 95th percentile under a normal approximation) as significant. The significant bigrams are collapsed into single tokens in every comment that contained them (\eg the bigram {\em false flag} is collapsed into a single token {\em false\_flag}). This process is performed recursively so that n-gram phrases can be identified.

Finally, for each time period, we construct independent word embeddings using the Word2Vec \cite{mikolov2013efficient} architecture based on the processed comments from that time period. Specifically, we train a continuous bag-of-words (CBOW) model with an embedding dimensionality of 100, a context window size of 5, and a minimum token frequency threshold of 5. Thus, each embedding captures the semantic relationships between words (and identified phrases) as they were used within the specific temporal context, independently of how they occurs in other time periods. We chose Word2Vec over more recent transformer-based models because \cite{basile2020diacr} found that diachronic embeddings perform better with static embeddings.

These embeddings serves as the basis for both RQ1 and RQ2. In RQ1, we use them to identify and validate conspiracy-related semantic regions. In RQ2, we align these embeddings across periods and analyze how the semantic neighborhoods associated with each conspiracy theory evolves over time.
\section{RQ1: Is Conspiratorial Language Semantically Distinguishable?} \label{sec:rq1}

Here, we ask whether the language associated with conspiracy theories occupies a coherent and semantically distinguishable region within all online political discourse.
Establishing this property is a prerequisite for studying the semantic evolution of our 19 conspiracy theories because without the presence of a well-defined semantic structure, individual conspiracy theories cannot be meaningfully represented, compared, or tracked over time.
To answer RQ1, we examine whether conspiracy-related discourse forms an internally coherent and externally separable region in the semantic space associated with all political discourse.
RQ1 may yield a negative result. If conspiratorial discourse is semantically indistinguishable from all other political discourse, clustering around our concept labels would show low coherence and human validation does not align with measures of semantic distance.
If either of these conditions occurs, we cannot identify semantic neighborhoods that represent individual conspiracy theories. 
Conversely, if we find coherent clusters around our concept labels and manually validated measures of semantic distance, this indicates that conspiratorial language is semantically distinguishable from other political discourse, and the semantic objects (clusters) around each concept label represent narratives surrounding each conspiracy theory.

\subsection{Methods}\label{sec:rq1:method}
Our goal is to determine whether each conspiracy theory occupies a localized and coherent region of the semantic space, as opposed to a diffuse or arbitrary collection of terms.
Intuitively, if a conspiracy theory has a meaningful presence in discourse, the words and phrases used to discuss it should cluster together in the semantic space.
To identify such regions, we treat each conspiracy theory's concept label as an anchor point and then ask: {\em what is the smallest semantic neighborhood that naturally forms around this label?} 
Rather than predefining the size of this neighborhood, we allow the structure of the embedding space to determine natural boundaries. 
If the natural boundary results in empty clusters (\ie only containing the concept label) or incoherent clusters (\ie containing terms unrelated to the conspiracy), then the semantic neighborhood is not distinguishable from other discourse.
On the other hand, coherent clusters would indicate a well-defined semantic structure that can be distinguished from other discourse.

\para{Identifying semantic neighborhoods.}
We operationalized the above intuition using density-based clustering. Specifically, we applied HDBSCAN \cite{mcinnes2017hdbscan} to identify clusters of high density in the semantic space.
We chose it because it is unsupervised, permits a variety of cluster shapes and sizes, and explicitly labels low-density regions as noise which provides a principled way to identify when a coherent semantic region does not exist.
%
%
We first applied it to the entire embedding space to identify all regions of meaningful semantic density as clusters.
Then, for each concept label, we located the cluster in which they appear and recursively used it to identify sub-clusters within their parent cluster. 
In effect, we looked for tighter and more coherent sub-clusters around the concept label in each iteration --- \ie in each iteration, peripheral and loosely connected terms were removed and only terms that consistently co-occurred with the concept label were maintained. 
We repeated this process until further refinement resulted in either an empty cluster or no change in the cluster boundary, which we used as an indication that a natural semantic boundary around the concept label had been achieved.
The final cluster around a concept label represents the semantic neighborhood associated with a given conspiracy theory.
Importantly, these neighborhoods are not defined by a fixed radius, nearest-neighbor counts, or keywords. Rather, they are defined by the existence of high-density semantic structures within the embedding space. 
Thus, if a conspiracy theory's concept label does not correspond to a coherent region, this procedure would fail to return a non-empty cluster. 
However, convergence would allow us to treat the final neighborhood (cluster) as a distinct semantic object.
By applying this process independently for each concept label, we obtain a set of concept-specific semantic regions that together define the conspiracy space in online political discourse (For additional implementation details, see Appendix \ref{sec:appendix}).

\begin{table*}[t]
\centering
\small
\begin{tabularx}{\textwidth}{l l X}
\toprule
{\bf Category} & {\bf Concept Label} &  {\bf Conspiracy Description } \\
                &  & [\textit{samples from semantic neighborhood}]\\
\midrule

\multirow{12}{*}{{\bf Staged Violence}}
& Crisis actor     & Victims and survivors of violent events are hired actors. \\
&   &   [\textit{hoaxer, paid actor}] \\
& False flag       & Acts committed with the intent to pin blame on another party. \\
&   &   [\textit{Soros paid, inside job}]\\
& Sandy Hook       & The school shooting was staged to advance gun control. \\
&   &   [\textit{push gun control, gun control advocate}]\\
& Boston bombing   & The bombing was orchestrated by the FBI. \\
&   &   [\textit{Boston marathon, terror attack}]\\
& Sutherland Springs       & The church shooting was staged by the Democratic party. \\
&   &   [\textit{Planned Parenthood shooting, Texas Church shooting}]\\
& Truther          & The 9/11 attacks were coordinated or permitted by the US government. \\
&   &   [\textit{inside job, 9/11}]\\
\midrule
\multirow{12}{*}{{\bf Political Scandals}}
& Emailgate        & Hillary Clinton's email server contained evidence of large-scale criminal acts. \\
&   &   [\textit{Whitewater, Clinton Foundation}]\\
& Benghazi         & The Benghazi attack involved deliberate inaction by Hillary Clinton. \\
&   &   [\textit{CIA operation, Hillary}]\\
& Russiagate       & Russian election interference was fabricated to undermine President Trump. \\
&   &   [\textit{debunked story, totally fake}]\\
& Birther          & President Obama was not born in the United States. \\
&   &   [\textit{whisper campaign, birtherism}]\\
& Voter fraud      & Elections were systematically manipulated through large-scale voter fraud. \\
&   &   [\textit{hacking vote machine, vote changed}]\\
& Death panels     & Public healthcare will result in bureaucrats determining treatment options. \\
&   &   [\textit{Obamacare, socialized medicine}]\\
\midrule
\multirow{10}{*}{{\bf Elite Control}}
& Illuminati       & A group of global elites control political and economic institutions. \\
&   &   [\textit{new world order, Bilderberg}]\\
& Deepstate        & Unelected intelligence officials control the government. \\
&   &   [\textit{indoctrination center, evil socialist}]\\
& Pizzagate        & Political elites operate underground child trafficking rings. \\
&   &   [\textit{Qanon, Seth Rich}]\\
& Lizard people    & Shape-shifting Lizards (or aliens) control the government. \\
&   &   [\textit{Illuminati, Reptilian}]\\
& Gamergate        & Video game developers and journalists coordinated to manipulate cultural norms in gaming. \\
&   &   [\textit{Kotaku in action, game journalism}]\\
\midrule
\multirow{4}{*}{{\bf Science Denial}}
& Chemtrails       & Contrails are biological agents sprayed by governments to control the populace. \\
&   &   [\textit{mind control, fluoride water}]\\
& Flat Earth       & The Earth is flat.\\
&   &   [\textit{Moon landing, round Earth}]\\
\bottomrule
\end{tabularx}
\caption{Overview of conspiracies, their concept labels, and sample words/phrases from their identified neighborhoods.}
\label{tab:rq1:labels}
\end{table*}

\subsection{Results} \label{sec:rq1:results}

\begin{figure}[!ht]
  \centering
  \includegraphics[width=3in]{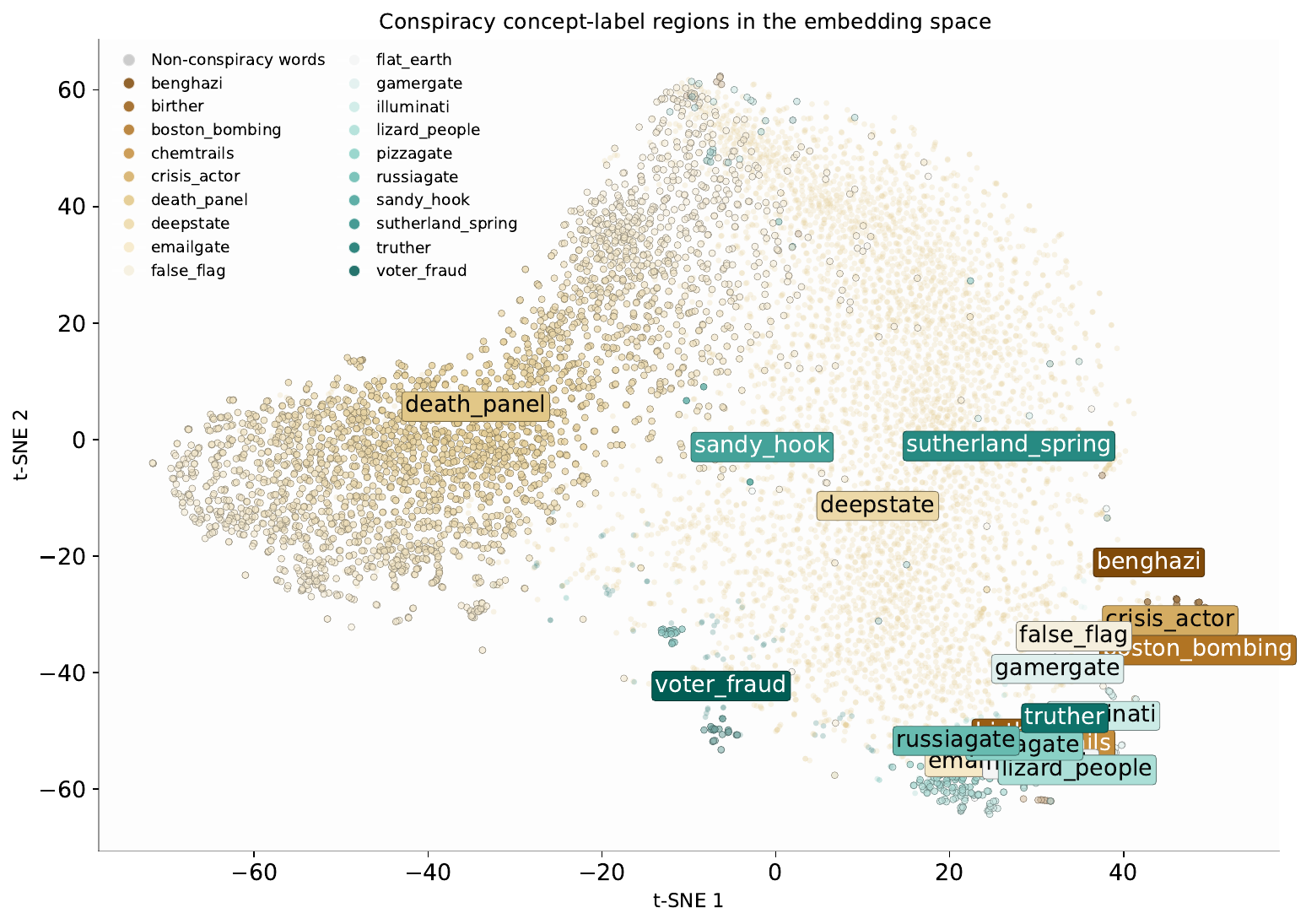}
  \caption{t-SNE projection of Word2Vec vocabulary highlighting conspiracy concept regions}
  \label{fig:conspiracies}
\end{figure}

\para{Evaluation of conspiracy neighborhood (cluster) overlaps and convergence.}
The t-SNE map (Fig. \ref{fig:conspiracies}) shows that conspiracy-related vocabulary forms coherent semantic neighborhoods within the broader embedding space, rather than being randomly dispersed among background terms. Applying the above procedure to all the 19 concept labels yielded 19 distinct semantic regions. Across the concepts, we observed strong internal coherence with only minor overlaps across similar categories of conspiracies (\eg Shooting-related conspiracies often overlapped). 
The final neighborhood sizes ranged from a few dozen terms to hundreds of terms, varying based on the amount of discourse associated with the conspiracy's concept label (more details in Appendix \ref{sec:appendix}).
Importantly, we found that \textit{all 19 concepts converged to a stable non-empty neighborhood}. 

\para{Qualitative evaluation of semantic neighborhoods.}
Qualitative analysis showed that \textit{clusters generally captured meaningful narratives that differed from generic political discourse or from other unrelated conspiracies}.
For example, the semantic neighborhood extracted for the `illuminati' concept label centered on themes of control by elites, including terms such as NWO (New World Order) and Bilderberg Group. In contrast, the `emailgate' concept label was anchored to themes of political mistrust, including terms such as Clinton corruption and numerous references to political scandals involving Hillary Clinton.
Both these regions are distinct from each other, as well as from other non-conspiratorial discourse. 
\Cref{tab:rq1:labels} offers canonical examples of the terms in the neighborhoods of each concept label.

\para{Experimental evaluation of alignment between semantic distance and human judgment.} 
Identifying semantic regions is insufficient on its own to establish that conspiracy-related discourse is meaningfully distinguishable from non-conspiratorial discourse. 
To validate this distinction, we conducted a human annotation study designed to test whether semantic distance from the conspiracy space corresponds to human judgments of conspiratorial meaning.
We computed, for every token in the non-conspiratorial embedding space, its minimum cosine distance to any concept label's neighborhood (cluster) centroid. This produced a distribution of distances reflecting proximity to the conspiracy space.
We partitioned this cosine similarity distribution into three distance-based levels using the mean and standard deviation of the distribution, with each level spanning one standard deviation. Words closer to the conspiracy space (the lower levels) are expected to be more difficult to distinguish from conspiratorial language, while farther away words (the higher levels) should be easier to distinguish.
Three annotators were presented with 1500 randomly chosen pairs of terms/phrases. Each pair included one term/phrase from the neighborhood of a conspiracy concept label and one term/phrase from the non-conspiratorial embedding space, drawn evenly from each distance level.
Annotators then had to identify which of the pair of words was associated with any of the 19 conspiracy theories.
To aid their decision making, they were given access to the descriptions of the 19 conspiracy theories in this study, as well as sample of {three} comments containing the concept label terms of each theory. 
This task allows us to measure whether the constructed semantic space effectively separates conspiratorial language from non-conspiratorial language, as determined by human judgment. If this were the case, we would expect the annotator's accuracy to show a clear gradient, highest in the pairs furthest from the conspiracy clusters and decreasing as the pairs get closer.
Our analysis of the annotator's accuracy, broken down by distance levels, showed precisely this trend --- \ie annotator accuracy increased as the semantic distance from the conspiracy space increased. 
To verify whether this trend was statistically significant, we conducted stratified bootstrapping by sampling 10K samples of 1500 word pairs (with replacement) while ensuring even representation from each annotator and each level in every sample. 
Computing the average accuracy for each level across all the bootstrapped samples, we found that the differences in accuracy between levels were statistically significant. 
Notably, the accuracy of pairs containing words from the farthest level (level 3) exceeded the accuracy of pairs containing words from the nearest level (level 1) by 12.64 percentage points. Smaller, but still significant differences were found for intermediate levels.
\Cref{tab:rq1:eval} illustrates the complete average annotator accuracy per distance level, as well as the difference in the mean accuracy for each level and 95\% confidence intervals obtained from our bootstrapped samples.
This alignment between semantic distance and human judgment indicates that the identified conspiracy space does in fact differ from non-conspiratorial language.

\begin{table}[t]
\centering
\small
\begin{tabular}{p{.75in} c c c}
\toprule
 & {\bf Level 1} & {\bf Level 2} & {\bf Level 3} \\
 & (Closest)     &               & (Farthest)   \\
\midrule
\multicolumn{4}{c}{\textit{Average accuracy across all annotators}} \\
\midrule
{ Accuracy (\%)} & 60.8\% & 70.4\% & 72.9\% \\
\midrule
\multicolumn{4}{c}{\textit{Bootstrap comparisons computed relative to Level~3 (farthest)}} \\
\midrule
{ $\Delta_{L3}$} & -12.6\%  & -4.0\% & -- \\
{ 95\% CI$_{\Delta L3}$} & [$-15.9$, $-9.3$] & [$-7.3$, $-0.8$] & -- \\
\bottomrule
\end{tabular}
\caption{Average annotator accuracy broken down by semantic distance levels from the conspiracy space and bootstrap-based 95\% confidence intervals for differences in mean annotator accuracy at each level relative to Level~3 (farthest). The differences in annotator accuracy were statistically significant across all three levels.}
\label{tab:rq1:eval}
\end{table}

\para{Takeaway.} Taken together, our clusters, qualitative, and experimental analysis demonstrate that conspiracy-related discourse forms coherent and semantically distinguishable regions within online political discourse.
Specifically, the cluster analysis confirms that conspiracy neighborhoods generally converge to no-to-low overlapping non-empty clusters, the qualitative analysis confirms the internal coherence of each cluster, and the experimental validation confirms the semantic distinguishability of conspiracy-related language from non-conspiratorial discourse.
These findings allow us to use the semantic neighborhoods (clusters) identified for each concept label as the {\em semantic object} representing the specific conspiracy theory.
In RQ2, we will use these semantic objects to track the evolutionary patterns of each conspiracy theory across the three time periods in our study.

\section{RQ2: How Do Conspiracies Evolve?} \label{sec:rq2}

Our analysis from RQ1 established that conspiracy-related discourse forms coherent and semantically distinguishable neighborhoods in the language space, which we can treat as {\em semantic objects} anchored on conspiracy concept labels.
These semantic objects reflect how a conspiracy theory is contextualized through its surrounding vocabulary at a given point in time. However, identifying these objects at a single time period does not explain how conspiracy theories change, persist, or transform.
Therefore, we now ask: {\em How do the semantic objects associated with conspiracy theories evolve over time?}

\subsection{Methods} \label{sec:rq2:methods}
\para{Constructing a temporal framework.}
To answer RQ2, we study the temporal changes in the semantics and lexica of the conspiracy-related semantic objects across three different time periods: 2012-2014 ($T_1$), 2015-2019 ($T_2$), and 2020-2022 ($T_3$) which reflect distinct social and political environments. 
$T_1$ reflects a period when conspiracy theories circulated at the fringes of political discourse, $T_2$ reflects a period during which conspiracy theories became more mainstream and politically salient, and $T_3$ reflects a period marked by the emergence of pandemic-related conspiracy theories.
We create three independently trained embedding spaces as described in \Cref{sec:dataset} and focus on how the semantic objects associated with our 19 conspiracy theories evolved from $T_1 \rightarrow T_2$ and $T_2 \rightarrow T_3$.

\para{Constructing real and shadow semantic objects.}
For each transition $T_i \rightarrow T_{i+1}$, we locate three semantic objects for each concept label ($l$):
(1) the semantic object of the concept label $l$ in $T_i$, which we refer to as $n_i(l)$;
(2) the semantic object of the concept label $l$ in $T_{i+1}$, which we refer to as $n_{i+1}(l)$; and
(3) the semantic object of the concept label $l$ that is created by projecting the $T_i$ embedding space into the $T_{i+1}$ embedding space using orthogonal Procrustes alignment \cite{schonemann1966generalized}, which we refer to as $n_{i \rightarrow i+1}(l)$.
Conceptually, $n_{i \rightarrow i+1}(l)$ provides the counterfactual baseline by capturing how the concept label's original meaning would appear if expressed in the subsequent time period without any semantic changes.
Divergence between $n_{i \rightarrow i+1}(l)$ and $n_{i+1}(l)$ would indicate that the meaning of the conspiracy theory associated with the concept label $l$ has evolved.

\para{Quantifying semantic evolution.} We compute three metrics to characterize the semantic evolution of a conspiracy.

\begin{enumerate}
    \item {\em Historical Association Divergence (HAD).} 
    This metric captures the extent to which the semantic associations that historically defined the conspiracy theory are preserved in the subsequent time periods. 
    Put differently, it measures whether the terms that used to define a specific conspiracy theory still relate to it with the same strength.
    For each concept label $l$, let the set $N(i,l)$ be the terms in in $n_i(l)$. Then, for each term $t \in N(i, l)$, we compute the cosine similarity between $l$ and $t$ in the embedding spaces associated with $T_i$ and $T_{i+1}$.
    We then compare these similarity distributions (\ie the distribution of similarities within $T_i$ with the distribution of similarities within $T_{i+1}$) using the Jensen-Shannon (JS) divergence \cite{lin2002divergence}.
    We use the JS divergence metric to capture the overall change in the concept label's association with the historical terms.
    To interpret the obtained JS divergence score, we calculate their $z$-scores using a baseline of JS divergence scores obtained from neighborhoods of randomly selected non-conspiracy terms.
    High HAD scores indicate that the terms that once jointly gave meaning to the conspiracy in $T_i$ no longer provide it the same support --- \ie the semantic associations that defined the conspiracy in $T_i$ are weaker in the $T_{i+1}$ time period.

    \item {\em Counterfactual Semantic Divergence (CSD).} 
    This metric captures the divergence between the counterfactual historic meaning and the observed contemporary meaning of the conspiracy. 
    Put differently, it measures how semantically different the current meaning of the conspiracy is from the historic meaning of the conspiracy. 
    Specifically, we compare the projected semantic neighborhood ($n_{i \rightarrow i+1}(l)$) with the observed neighborhood $n_{i+1}(l)$ by computing the average pairwise cosine distance between all terms across the two neighborhoods.
    We use this approach as it effectively measures how far the counterfactual historic neighborhood lies from the observed contemporary neighborhood.
    To interpret the obtained average pairwise cosine distance, we calculate their $z$-scores using a baseline of average cosine distances obtained using neighborhoods of randomly selected non-conspiracy terms.
    High CSD values indicate that the projected historic meaning lies far from the observed contemporary meaning, providing evidence of significant semantic change rather than surface-level semantic drift.
    Note that it is possible for a conspiracy to maintain strong historical associations (low HAD score), but still reflect a significantly different meaning (high CSD score).
    
    \item {\em Neighborhood Coherence Divergence (NCD).} 
    This metric captures whether a conspiracy concept label has gone from expressing one coherent meaning to expressing multiple semantically distinct meanings over time.
    Put differently, it measures if the terms associated with the conspiracy shifted from one unified semantic structure to multiple internally coherent semantic structures.
    Unlike CSD and HAD, which measure aggregate semantic displacement and loss of historical associations respectively, NCD is designed to detect changes in the {internal structure} of a concept’s semantic neighborhood.
    For each concept label $l$, we construct a fixed union set of terms: $U(i,l) = n_i(l) \cup n_{i+1}(l)$
    ensuring that both historically salient and newly emergent semantic contexts are evaluated jointly.
    We then extract the embeddings of all terms $t \in U(i,l)$ from the embedding spaces associated with $T_i$ and $T_{i+1}$, yielding two sets of vectors, $V^{i}$ and $V^{i+1}$.
    For each vector set, we compute all pairwise cosine similarities and calculate the kurtosis \cite{pearson1905fehlergesetz} of the resulting similarity distribution.
    Kurtosis captures the extent to which the similarity distribution exhibits heavy tails, which arise when the neighborhood contains tightly cohesive subgroups with weak similarity between them.
    We define NCD as the change in kurtosis between $T_i$ and $T_{i+1}$.
    To interpret this change across concept labels, we standardize the observed kurtosis difference using a baseline distribution obtained from randomly selected non-conspiracy neighborhoods, yielding a $z$-score.
    Large positive NCD values indicate that the neighborhood has become more internally polarized over time, consistent with the emergence of additional semantic structure.
    Conversely, large negative NCD values indicate that the neighborhood has become more internally coherent and  centered around a single meaning.    
\end{enumerate}

Together, CSD, HAD, and NCD capture complementary aspects of semantic evolution. 
CSD measures whether the historical meaning of a conspiracy theory has shifted to a different semantic location, HAD measures whether the historical associations that once defined the theory remain intact, and NCD measures whether the internal structure of the theory’s semantic neighborhood remains coherent or reorganizes into multiple distinct meanings. 
Interpreted jointly, these metrics distinguish semantic displacement, association erosion, and internal fragmentation that reflect three distinct evolutionary patterns.

\begin{table}[t]
\centering
\small
\begin{tabular}{l c c c c}
\toprule
{\bf Pattern} & {\bf HAD} & {\bf CSD} & {\bf NCD} & {\bf LO} \\
\midrule
Sem. Stability & Low & Low & Low & High\\
Sem. Narrowing & High & Low & Low & * \\
Sem. Replacement & High & High & * & * \\
Sem. (De)fragmentation & * & * & High & High \\
Lex. Replacement & Low & Low & Low & Low \\
\bottomrule
\end{tabular}
\caption{Rules for assigning conspiracy theories to evolutionary patterns based on semantic and lexical metrics. A '*' indicates that the metric may take any value.}
\label{tab:rq2:rules}
\end{table}

\para{Quantifying lexical evolution.}
While the above metrics characterize how the meaning of a conspiracy theory evolves, conspiracies may also change through shifts in the vocabulary used to express them, regardless of whether their meaning changes.
To capture this dimension, we also quantify lexical changes independently of semantic evolution.
For each concept label $l$ associated with a conspiracy, we measure lexical change by comparing the vocabulary of the semantic neighborhoods $n_i(l)$ and $n_{i+1}(l)$. 
Specifically, we compute the Jaccard similarity between the two neighborhood vocabularies. 
A low similarity indicates a high turnover in vocabulary and a high similarity indicates lexical stability.
To interpret the observed similarities for each concept, we calculate their $z$-scores using a baseline of Jaccard similarities obtained using neighborhoods of randomly selected non-conspiracy terms.
A high $z$-score indicates that the lexical turnover for a conspiracy theory is substantially higher than the average lexical turnover for non-conspiracy theories.
We refer to this metric as the {\em Lexical Overlap (LO).}

\para{Characterizing evolutionary patterns.}
We characterize the evolutionary patterns of each conspiracy theory by interpreting their semantic and lexical metrics across $T_1$, $T_2$, and $T_3$ as follows. \Cref{tab:rq2:rules} summarizes the patterns.

\begin{itemize}
    \item {\em Semantic stability.} A conspiracy theory is classified as exhibiting semantic stability between two time periods when all three semantic metrics (HAD, CSD, and NCD) remain low ($z$-score < 1).
    In this case, the historical associations from the older time period persist, the counterfactual historic meaning aligns with the contemporary meaning, and the internal structure remains coherent.
    Together, these indicate that neither the meaning or structure of the theory has changed in a meaningful way.

    \item {\em Semantic replacement.} A conspiracy theory is classified as undergoing semantic replacement when both HAD and CSD are high ($z$-score > 1). 
    High HAD indicates that the historical associations that once defined the conspiracy have weakened or disappeared, while high CSD indicates that the projected historic meaning lies far from contemporary usage. 
    When these two conditions co-occur, they indicate that the concept label has come to represent a meaningfully different idea than previously.

    \item {\em Semantic narrowing.} A conspiracy theory is classified as broadening or narrowing when HAD is high, but CSD and NCD are low.
    Here, the high HAD indicates that some of the historical associations of the conspiracy have weakened or dropped out, but the low CSD indicates that the core meaning is still aligned with the past.
    Together, this indicates that the theory is being reshaped by dropping old associations and adding new ones.
    The unchanged NCD suggests that the theory remains as unified/fragmented as before.

    \item {\em Semantic fragmentation.} A conspiracy theory is identified as undergoing semantic fragmentation when NCD exhibits a large positive change ($z$-score > 1), while the HAD and CSD remain low ($z$-score < 1).
    Here, the elevated NCD indicates that the semantic neighborhood has multiple internally coherent structures, reflecting the existence of multiple meanings.

    \item {\em Semantic defragmentation.} A conspiracy theory is identified as undergoing semantic defragmentation when NCD exhibits a large negative change across time ($z$-score < -1), while the HAD and CSD remain low ($z$-score < 1)
    This pattern indicates that the internal structure has become more centralized and that previously loosely connected interpretations of the theory have consolidated into a single dominant meaning.

    \item {\em Lexical replacement.} A conspiracy theory exhibits lexical replacement when it is semantically stable, but the lexical overlap across time is low.
    This pattern indicates that the meaning of the conspiracy remains unchanged, but the vocabulary used to express it have changed significantly.
    
\end{itemize}

\para{Measuring relative semantic drift.}
In addition to studying individual conspiracy theories in isolation, we also examine how conspiracy theories evolve relative to each other --- \ie which other conspiracy theories do they become more or less closely associated with over time?
For each time period and conspiracy theory, we compute the cosine similarity between the centroids of the given conspiracy theory and all other conspiracy theories.
We then track the degree to which these similarities have increased or decreased in subsequent time periods.
If the similarity between two theories has increased, we characterize them as becoming more closely interlinked. Conversely, a decrease in similarity is associated with increased semantic divergence between the theories.

\begin{figure*}[t]
    \centering
    \begin{subfigure}[t]{0.48\textwidth}
        \centering
        \includegraphics[width=\linewidth]{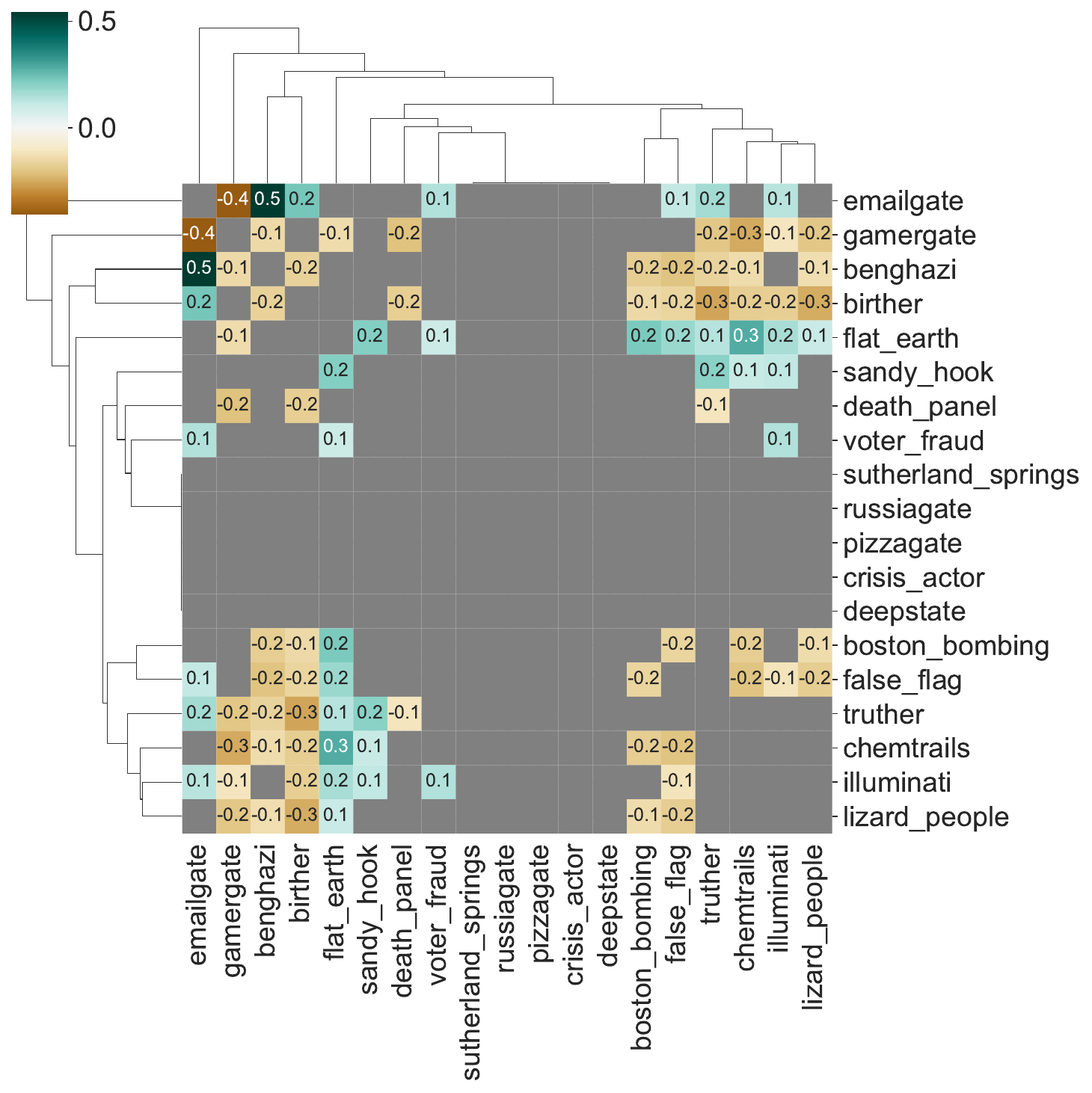}
        \caption{Change in pairwise conspiracy similarity from $T_1$ to $T_2$.}
        \label{fig:rq2:heatmap_t1_t2}
    \end{subfigure}
    \hfill
    \begin{subfigure}[t]{0.48\textwidth}
        \centering
        \includegraphics[width=\linewidth]{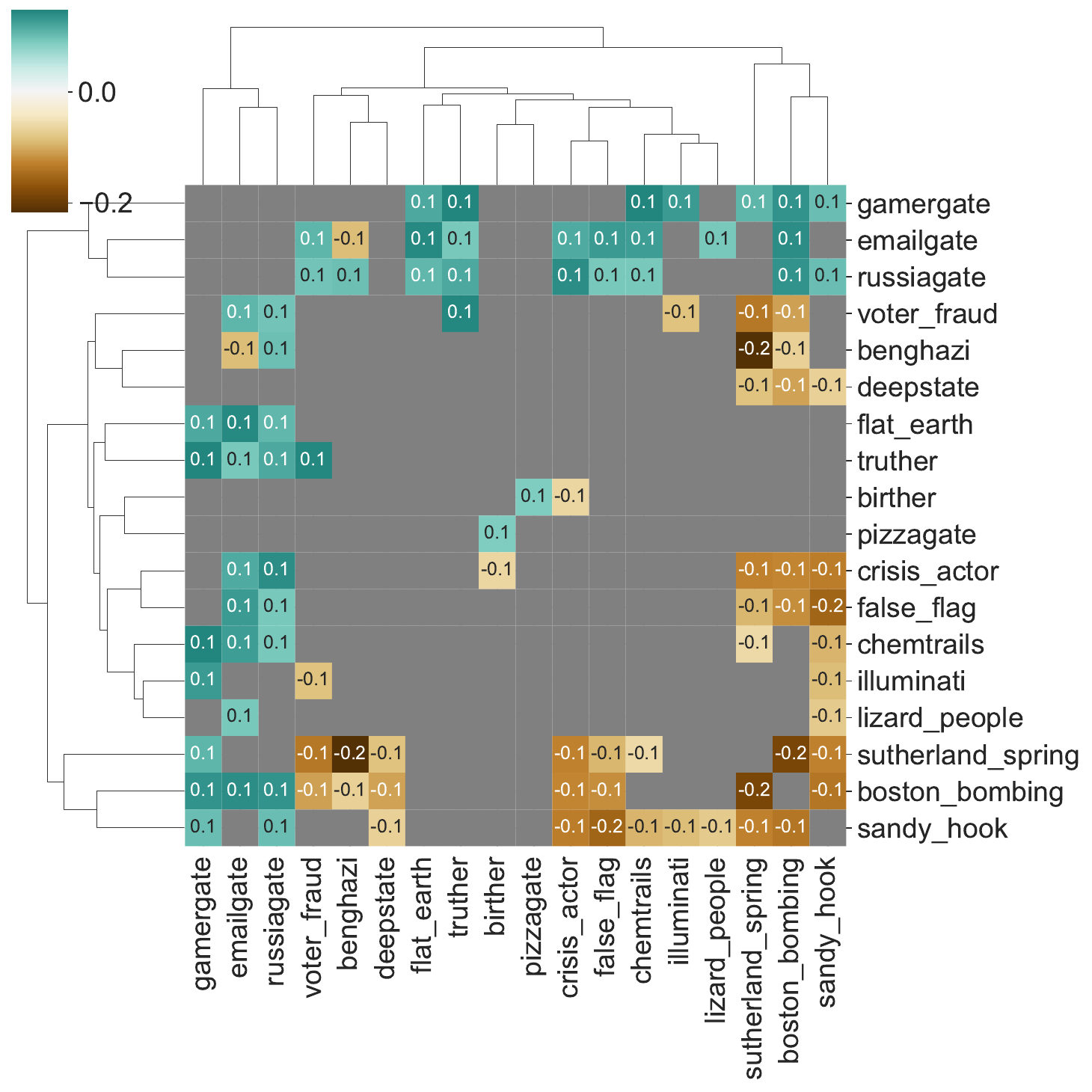}
        \caption{Change in pairwise conspiracy similarity from $T_2$ to $T_3$.}
        \label{fig:rq2:heatmap_t2_t3}
    \end{subfigure}
    \caption{Relational reorganization of the conspiracy landscape across time. Cell values indicate the change in semantic similarity between conspiracy theories, blue cells indicate convergence and brown cells indicate divergence. Only statistically significant changes are shown.}
    \label{fig:rq2:heatmap_comparison}
\end{figure*}

\subsection{Results} \label{sec:rq2:results}
We now examine how different conspiracy categories evolved through $T_1 \rightarrow T_2$ and $T_2 \rightarrow T_3$ by interpreting the evolution metrics described in \Cref{sec:rq2:methods}.

\para{Staged violence conspiracies exhibit high lexical churn while maintaining semantic stablity.} Conspiracies centered around claims that violent events were staged or fabricated (\eg Sandy Hook, Sutherland Springs, and Boston Bombing) are characterized by semantic stability and substantial lexical evolution.
Across both time period transitions, these conspiracies exhibited low HAD and CSD, indicating that their meanings remained aligned with historical usage and prior semantic associations.
Despite this, however, the conspiracies showed consistently low lexical overlap, indicating consistent churn in vocabulary.
Upon inspection of the vocabulary, we find that this is because conspiracies in this category appear to be event driven with nearly identical narratives.
Put differently, when a new mass casualty incident occurs, discourse simply incorporates new locations and actors while preserving the narratives of false flags, government motives, and crisis actors.
For example, `Sandy Hook' remained anchored in shooting denial narratives, but expanded lexically to include locations and actors involved in Parkland, Las Vegas, and other shooting events.
Examining the structure of the semantic neighborhoods, we find that several conspiracies in this category had strong negative NCD scores in the transition from $T_2 \rightarrow T_3$, indicating semantic defragmentation.
This suggests that discourse in the latter part of our study period consolidated around a smaller number of interpretive frames.
Overall, conspiracies associated with claims of staged violence evolved through lexical replacement and concept consolidation.

\para{Political scandal conspiracies underwent semantic replacement.}
Politically grounded conspiracy theories such as Emailgate, Benghazi, Russiagate, and death panels exhibited the most pronounced cases of semantic replacement, particularly in the $T_1 \rightarrow T_2$ transition.
This makes intuitive sense: the conspiracies are associated with specific institutions and actors, making their semantics more susceptible to change as political conditions evolve.
Of all the conspiracies in this category, `emailgate' serves as a canonical example of this behavior. 
Early discourse in $T_1$ around emailgate was dispersed across multiple political grievances before collapsing into a highly focused narrative centered on institutional corruption in $T_2$.
This semantic replacement was accompanied by semantic defragmentation, as reflected by large negative NCD scores, indicating consolidation around this single narrative.
Following this semantic replacement in $T_1 \rightarrow T_2$, political conspiracies saw little semantic change in $T_2 \rightarrow T_3$.
Instead, they exhibited lexical evolution by incorporating new actors (\eg Joe Biden and Hunter Biden) into existing narratives of institutional corruption.

\para{Elite control conspiracies exhibit high fragmentation.}
Conspiracies related to elite control over society (\eg illuminati and deep state) are characterized by persistently high NCD values during both transitions from $T_1 \rightarrow T_2$ and $T_2 \rightarrow T_3$.
These conspiracies retained a core meaning centered around secret coordination and institutional manipulation by hidden elites. However, they also attracted new interpretations over time.
For example, the `illuminati' conspiracy maintained its elite-control framing while incorporating new narratives related to satanic cabals and Qanon, reflecting a semantic neighborhood with multiple internally coherent meanings.
This pattern of high fragmentation while maintaining semantic relationships suggest that conspiracy theories within this class operate as meta-conspiracies which provide a flexible interpretive lens to accommodate new narratives without replacing prior ones.

\para{Science denialism is anchored by a stable epistemic stance.}
Conspiracy theories rooted in science denial exhibit remarkably stable semantics (low HAD and CSD) across both time period transitions, indicating that their core narrartives around the rejection of scientific authority remain intact over time.
At the same time, NCD increases during the transition from $T_1 \rightarrow T_2$, reflecting the emergence of multiple coherent sub-narratives. 
In the case of the flat earth conspiracy, early discourse was grounded in narratives around creationism and debates around science and religion. This grew to expand narratives around faked scientific evidence and scientific hoaxes, before later absorbing 5G and anti-vax narratives.
Similar trends were also observed for the `chemtrails' conspiracy which evolved from geoengineering-focused narratives to conspiracies about vaccine nanochips.
In $T_3$, the science denial-related conspiracies generally exhibited high defragmentation as the narratives around the pandemic and vaccines became dominant.

\para{The conspiracy landscape experiences relative evolution.}
To complement the per-conspiracy analysis of evolution, we also examined how the relative positions of each theory changed with respect to one another across time.
\Cref{fig:rq2:heatmap_comparison} shows the heatmaps of the pairwise changes in semantic similarity between conspiracy theories for the transitions $T_1 \rightarrow T_2$ and $T_2 \rightarrow T_3$, respectively.
The blue cells indicate that the conspiracies have become more semantically similar in the latter time period, while the brown cells indicate a reduced similarity. Only changes that are higher than one standard deviation are shown.
In the transition from $T_1 \rightarrow T_2$, we find that the changes are concentrated among specific groups of conspiracies.
The political scandals (particularly emailgate and Benghazi) show strong convergence, confirming consolidation into a coherent cluster associated with institutional corruption in $T_2$.
We also find that after the US Presidential elections in 2016, emailgate and gamergate experienced a sharp semantic divergence.
Interestingly, the staged violence group of conspiracy theories experience a different trajectory as they increasingly decouple themselves from political scandals in $T_2$ and begin to converge among themselves in $T_3$.
This pattern supports our earlier finding of semantic stability with lexical churn.
Conspiracy theories in the science denial and elite control groups move closer towards each other in $T_3$, indicating the development of a broader class of conspiracies based on institutional mistrust (\eg Qanon, deepstate, vaccine microchips, and 5G).
In general, the transition from $T_1 \rightarrow T_2$ highlights that conspiracies experienced significant reorganization and repositioning of conspiracies relative to each other. In contrast, the transition from $T_2 \rightarrow T_3$ is marked by consolidation and increased coherence within conspiracy classes.

\para{Takeaway.}
Our analysis reveals that conspiracy theories exhibit distinct evolutionary patterns based on their relationships with institutional authorities and political events.
Semantic replacement is a rare occurrence and appears concentrated in the political scandal category in the transition from $T_1 \rightarrow T_2$, when the flood of misinformation on social media led to the mainstreaming of conspiratorial discourse.
More commonly, in this period, conspiracies appeared to adapt through semantic fragmentation. 
In the later transition from $T_2 \rightarrow T_3$, we found that no class of conspiracies experienced any meaningful semantic changes. 
Instead, nearly all conspiracies experienced consolidation and refinement, with previously fragmented meanings becoming more coherent around a smaller number of narratives.

\section{Related Work} \label{sec:related-work}

\para{Semantic evolution.}
``You shall know a word by the company it keeps'' \cite{firth1957synopsis}. But the company is never static, it shifts at different rates for different words \cite{blank1999new}.
%
A standard way to understand this evolution is to compare word embeddings across different periods of time \cite{jatowt2014framework, kulkarni2015statistically, hamilton2016diachronic, basile2020diacr, giulianelli2023interpretable}.
Kulkarni et al. (\citeyear{kulkarni2015statistically}) is one of the earlier works that treats semantic change as a time series problem to detect \textit{when} has the shift occurred, finding that a shift in meaning is often preceded by a shift in frequency or syntactic usage (part-of-speech).
Similarly, Hamilton et al. (\citeyear{hamilton2016diachronic}) also did foundational work in exploring semantic change by analyzing six historical corpora spanning four languages (English, German, French, and Chinese). Using Orthogonal Procrustes analysis to align embeddings (PPMI, SVD, and word2vec) they posited that frequent words change at slower rates, and polysemous words are inherently more prone to semantic evolution, independent of their frequency.

With the rise in transformer-based models, Basile et al. (\citeyear{basile2020diacr}) tailored a shared task (DIACR-Ita) to detect automatic semantic shift across different models. Their findings showed that traditional static embeddings (e.g., SGNS aligned with Procrustes) often outperformed contextualized models for binary change detection tasks.
Giulianelli et al. (\citeyear{giulianelli2023interpretable}) move beyond simply ``detecting'' the shift to the ``explaining'' why it occurred. By using Large Language Models (e.g., Flan-T5) to generate natural language definitions for word usages, they construct \textit{Sense Dynamics Maps} that explicitly describe how and why a word changed (e.g., narrowing, broadening) rather than just providing a numerical change score.

A major critique of standard diachronic shift methods is that they rely on aligning separate embeddings for each time period, which makes them prone to error and instability.
Gonen et al. (\citeyear{gonen2020simple}) demonstrate this instability by changing random seeds or hyperparameters, drastically altering the list of ``changed'' words as a result. They propose a stable alternative that bypasses alignment entirely by directly comparing the intersection of a word's nearest neighbors across different time stamps.
Similarly, Periti et al. (\citeyear{periti2025studying}) devise another approach which bypasses the need to align by treating the corpus as a dynamic stream. Their model retains a ``memory'' of the previous observed data, allowing addition of newer data without alignment.

While earlier literature primarily focused on detecting \textit{when} words shift meaning, recent scholarship has pivoted toward understanding \textit{why} and \textit{how} these shifts occur. Our work distinguishes itself from this prior research in two key respects. First, rather than analyzing isolated word-level shifts, we model the evolution of concept spaces -- defined as coherent semantic neighborhoods anchored to specific conspiratorial themes. This allows us to trace the systemic evolution of narratives. Second, we explicitly decouple semantic evolution from lexical churn. By independently measuring changes in meaning (via embedding alignment) and changes in vocabulary (via lexical overlap), we identify evolutionary patterns -- such as semantic stability amidst lexical churn -- that purely keyword-based approaches fail to capture.

\para{Online conspiratorial space.} Environments of low confidence and low trust serve as a breeding ground for conspiracies \cite{shahsavari2020conspiracy}. Such spaces when in the presence of exclusive support for similar opinions and restricted opposing view points increase the susceptibility of adapting conspiratorial thinking \cite{koutra2015events}. Bessi et al. (\citeyear{bessi2015trend}) observed that users tend to aggregate in highly polarized communities (echo chambers) where information is consumed to reinforce existing beliefs rather than to verify facts. This structural isolation provides the necessary environment for distinct semantic neighborhoods to form and evolve independently of mainstream correction \cite{starbird2017examining, koutra2015events}.

Within these isolated spaces, communities establish unique identities and structures. Khalid et al. (\citeyear{khalid2020style}) found that online communities (such as r/politics) establish unique stylistic fingerprints -- defined by functional markers rather than content -- which is indicative of a sociolinguistic identity. Furthermore, large-scale studies have demonstrated that this content is not random noise but organizes around recurring thematic domains such as environment, diet, geopolitics, and health \cite{bessi2015trend}. 
However, the structural topology is not uniform. \cite{mahl2021nasa} mapped the communication networks of 10 prominent conspiracy theories on Twitter, revealing a complex landscape where some narratives (e.g., QAnon and Pizzagate) merge into interconnected super-conspiracies, while others (e.g., Flat Earth) remain in distinct, isolated sub-communities.
%
Further, within these communities, conspiracy theories share a stable ``narrative grammar''. \cite{samory2018government} identify consistent ``agent-action-target'' motifs (e.g., Agency X deceives Population Y) that recur across range of topics, suggesting that the semantic core of a conspiracy often relies on abstract relational structures rather than specific vocabulary. 
While these conspiratorial spaces possess a distinct structure and narrative grammar, they are far from static. Rather, they function as dynamic, event-driven environments. Starbird (\citeyear{starbird2017examining}) found that an ecosystem of alternative media domains sustains these narratives by rapidly repurposing existing tropes (such as `false flag' or `crisis actor') for new events. This network effectively launders fringe claims into broader discourse. These findings are similar to ours (see section \cref{sec:rq2:results}) where the lexical churn is high but the conspiracy is semantically stable, specifically for the words false flag, government motive, and crisis actors.
%
%

While most prior literature studies structure, propagation, and semantics of a single snapshot, we do a longitudinal analysis from 2012-2022. Our approach distinguishes itself by decoupling the lexical and semantic dimensions of conspiratorial discourse, allowing us to identify holistic evolutionary patterns.
Furthermore, whereas previous studies often compare language across multiple communities, we restrict our analysis to a single mainstream community, r/politics. This focus ensures greater consistency in our comparisons throughout the study period.
\section{Discussion and Concluding Remarks} \label{sec:discussion}
In this work, we ask whether conspiratorial discourse occupies coherent and semantically distinguishable regions of language, and if so, how those regions evolve over time.
After establishing that conspiracy-related discourse can be captured by distinct semantic objects, we develop mechanisms to track how they change over time.
A key finding of our study is that semantic and lexical expressions can operate independently and that recognizing this distinction helps explain how conspiracies evolve.

\para{Implications for understanding conspiracy-related discourse.}
Our analysis shows that complete semantic replacement of conspiracies is rare and that the fundamental ideas underlying many conspiracy theories remain stable, even as their vocabulary evolves.
Similar to prior work, we find that event-centric conspiracies tend to exhibit primarily lexical changes by way of updating the actors and locations, while preserving their semantic framing.
Other conspiracies evolve by fragmenting into multiple distinct ideas or consolidating around a smaller number of ideas.
These patterns show the processes through which conspiratorial narratives adapt linguistically to changing environments, without requiring their subscribers to abandon their prior beliefs.
The distinction between semantics and vocabulary that are highlighted in this study is important: approaches that rely solely on vocabulary risk misclassifying stable conspiracies as newly emerging because of conflating lexical and semantic change. 
By considering theories as semantic objects rather than a collection of keywords, our methods provide a more faithful account of how conspiratorial discourse persists, adapts, and restructures over time.

\para{Implications for social media platforms and users.}
Prior work has shown that once individuals and communities engage with conspiratorial content, repeated exposure can deepen and reinforce their beliefs \cite{10.1145/3555551, Habib_Nithyanand_2022}.
Our methods allow us to identify emerging conspiratorial semantic spaces as they form and evolve, rather than after they have consolidated.
When applied to real-time data, our framework can support key stakeholders such as journalists and platform moderators by flagging discourse that is beginning to consolidate into recognizable conspiratorial structures.
Although such earlier identification cannot eliminate conspiratorial belief formation, it enables proactive and nuanced interventions including fact checking or deplatforming before problematic narratives become widely normalized.

\para{Limitations.} This study is not without limitations. First, due to challenges with data availability, our analysis is limited to the decade between 2012 and 2022.
While this interval captures a range of conspiratorial narratives and major socio-political transitions, it does not reflect more recent developments post-2022.
Second, our analysis begins with a fixed set of 19 conspiracy concepts. Although these represent well-established and widely-discussed narratives, they do not exhaust the full space of political conspiracy discourse.
Finally, our study focuses on conspiratorial discourse within a single mainstream political community on Reddit. 
While this setting allows us to observe how conspiracy narratives are expressed in the political and mainstream context, evolutionary dynamics may differ in more fringe communities or those with different moderation protocols.

\section*{Ethical Statement} \label{sec:ethicalstat}

Our study used publicly available data through publicly released Pushshift dumps \cite{reddit_pushshift_dump}. We collected comment text and associated timestamps from the subreddit {\tt r/politics} within the period between 2012 and 2022. No private messages, deleted content, or metadata beyond post dates and comment text were used. The study did not involve any direct interaction with Reddit users, nor did it attempt to infer or analyze user identities or demographics.
\bibliography{aaai26}
\section*{Paper Checklist}

\begin{enumerate}

\item For most authors...
\begin{enumerate}
    \item  Would answering this research question advance science without violating social contracts, such as violating privacy norms, perpetuating unfair profiling, exacerbating the socio-economic divide, or implying disrespect to societies or cultures?
    \answerYes{Yes, see Ethical Considerations section}
  \item Do your main claims in the abstract and introduction accurately reflect the paper's contributions and scope?
    \answerYes{Yes}
   \item Do you clarify how the proposed methodological approach is appropriate for the claims made? 
    \answerYes{Yes, please refer to the Methodology section}
   \item Do you clarify what are possible artifacts in the data used, given population-specific distributions?
    \answerYes{Yes, we discuss that in Limitations section}
  \item Did you describe the limitations of your work?
    \answerYes{Yes, we do that in the Limitations section}
  \item Did you discuss any potential negative societal impacts of your work?
    \answerYes{Yes, we discuss the societal impact of our work in Discussion section}
      \item Did you discuss any potential misuse of your work?
    \answerTODO{No, our work discusses established conspiracy theories, if anything, we are spreading awareness about these to mitigate their widespread.}
    \item Did you describe steps taken to prevent or mitigate potential negative outcomes of the research, such as data and model documentation, data anonymization, responsible release, access control, and the reproducibility of findings?
    \answerYes{Yes,  our  data  does  not  include  any privacy  sensitive information, we have made the code/details of the code where needed to ensure reproducibility.}
  \item Have you read the ethics review guidelines and ensured that your paper conforms to them?
    \answerYes{Yes}
\end{enumerate}

\item Additionally, if your study involves hypotheses testing...
\begin{enumerate}
  \item Did you clearly state the assumptions underlying all theoretical results?
    \answerNA{NA}
  \item Have you provided justifications for all theoretical results?
    \answerNA{NA}
  \item Did you discuss competing hypotheses or theories that might challenge or complement your theoretical results?
    \answerNA{NA}
  \item Have you considered alternative mechanisms or explanations that might account for the same outcomes observed in your study?
    \answerNA{NA}
  \item Did you address potential biases or limitations in your theoretical framework?
    \answerNA{NA}
  \item Have you related your theoretical results to the existing literature in social science?
    \answerNA{NA}
  \item Did you discuss the implications of your theoretical results for policy, practice, or further research in the social science domain?
    \answerNA{NA}
\end{enumerate}

\item Additionally, if you are including theoretical proofs...
\begin{enumerate}
  \item Did you state the full set of assumptions of all theoretical results?
    \answerNA{NA}
	\item Did you include complete proofs of all theoretical results?
    \answerNA{NA}
\end{enumerate}

\item Additionally, if you ran machine learning experiments...
\begin{enumerate}
  \item Did you include the code, data, and instructions needed to reproduce the main experimental results (either in the supplemental material or as a URL)?
    \answerNA{NA}
  \item Did you specify all the training details (e.g., data splits, hyperparameters, how they were chosen)?
    \answerNA{NA}
     \item Did you report error bars (e.g., with respect to the random seed after running experiments multiple times)?
    \answerNA{NA}
	\item Did you include the total amount of compute and the type of resources used (e.g., type of GPUs, internal cluster, or cloud provider)?
    \answerNA{NA}
     \item Do you justify how the proposed evaluation is sufficient and appropriate to the claims made? 
    \answerNA{NA}
     \item Do you discuss what is ``the cost`` of misclassification and fault (in)tolerance?
    \answerNA{NA}
  
\end{enumerate}

\item Additionally, if you are using existing assets (e.g., code, data, models) or curating/releasing new assets, \textbf{without compromising anonymity}...
\begin{enumerate}
  \item If your work uses existing assets, did you cite the creators?
    \answerNA{NA}
  \item Did you mention the license of the assets?
    \answerNA{NA}
  \item Did you include any new assets in the supplemental material or as a URL?
    \answerNA{NA}
  \item Did you discuss whether and how consent was obtained from people whose data you're using/curating?
    \answerNA{NA}
  \item Did you discuss whether the data you are using/curating contains personally identifiable information or offensive content?
    \answerNA{NA}
\item If you are curating or releasing new datasets, did you discuss how you intend to make your datasets FAIR?
\answerNA{NA}
\item If you are curating or releasing new datasets, did you create a Datasheet for the Dataset? 
\answerNA{NA}
\end{enumerate}

\item Additionally, if you used crowdsourcing or conducted research with human subjects, \textbf{without compromising anonymity}...
\begin{enumerate}
  \item Did you include the full text of instructions given to participants and screenshots?
    \answerNA{NA}
  \item Did you describe any potential participant risks, with mentions of Institutional Review Board (IRB) approvals?
    \answerNA{NA}
  \item Did you include the estimated hourly wage paid to participants and the total amount spent on participant compensation?
    \answerNA{NA}
   \item Did you discuss how data is stored, shared, and deidentified?
   \answerNA{NA}
\end{enumerate}

\end{enumerate}

\appendix
\section{Appendix} \label{sec:appendix}

\textbf{Clustering implementation details.}
The clustering pipeline is designed to handle a large vocabulary of approximately 4M words derived from Word2Vec model. Since Word2Vec vectors are already meaningfully compressed into a relatively low-dimensional semantic space, there is no need to apply an additional dimensionality reduction like UMAP before clustering. The first stage is creating a centroid set. Because clustering 4 million vectors all at once with HDBSCAN would be computationally prohibitive, the pipeline begins by running HDBSCAN on a manageable subset of the vocabulary (the first 15,000 vectors) to get coarse cluster assignments. Once this initial clustering is done, we compute the centroid of each discovered cluster, producing one representative vector per cluster. 

The second stage extends cluster assignments to the remaining unlabeled words in the full 4M word vocabulary. For every word that was not included in the initial HDBSCAN run (i.e., words beyond the first 15,000), the pipeline computes the cosine distance between that word's vector and each cluster centroid, then assigns the word to the nearest centroid. The result is a full mapping of all vocabulary words to one of the initial clusters. 

The third stage is sub-clustering, which is applied to each concept label. Starting from the full set of words in the initial cluster containing the target term, the pipeline re-runs HDBSCAN on that subset. After HDBSCAN identifies sub-clusters within this sample, the centroids of those sub-clusters are computed, and then every word in the full (unsampled) cluster is re-assigned to its nearest sub-cluster centroid using cosine distance. This produces a refined, smaller cluster. The process then repeats on the sub-cluster containing the target term, drilling down level by level. The recursion terminates when HDBSCAN can no longer find any meaningful structure in the remaining words i.e. when every word is labeled as noise (-1), indicating that the cluster has been narrowed down to a tight, semantically coherent core neighborhood around the conspiracy term. The final cluster at each term's termination point is saved as that term's representative word neighborhood, capturing the semantic context known as semantic object. 

Table~\ref{tab:subclustering} presents sample results from the sub-clustering process for four concept labels. To illustrate how the recursive sub-clustering converges, we report the number of words present at each iteration level — starting from the broad initial cluster assigned and ending at the tightest neighborhood where HDBSCAN can no longer detect meaningful density structure. The ``Words Removed'' column gives a sample of words that were present in earlier iterations but fell outside the final cluster, showing how the process gradually removes loosely related words at each step, leaving behind only the words most closely associated with the target concept label.

\begin{table}[t]
\centering
\small
\setlength{\tabcolsep}{4pt}
\caption{Sample sub-clustering results for selected concept labels. 
Word counts at each iteration are shown in sequence; the final value denotes the 
final neighborhood size after recursive sub-clustering converges.}
\label{tab:subclustering}
\begin{tabularx}{\columnwidth}{p{1.8cm} c p{2.0cm} X}
\toprule
\textbf{Concept} & \textbf{Words per Level} & \textbf{Sample Words Removed} \\
\midrule

\texttt{sandy\_hook} & {[}10264, 252, 41, \textbf{30}{]} &
\textit{accidentally\_shot, james\_holmes, jared\_loughner, shooting\_since} \\
\addlinespace

\texttt{chemtrails} & {[}10418, 460, 294, \textbf{278}{]} &
\textit{canola, cask, coral\_reef, fished, lead\_based, sprayed\_roundup} \\
\addlinespace

\texttt{crisis\_actor} & {[}4298, 26, \textbf{16}{]} &
\textit{audio\_tape, feud, operation\_chaos} \\
\addlinespace

\texttt{birther} & {[}4298, 69, \textbf{39}{]} &
\textit{bullshit\_attack, called\_romney, crazy\_statement, mccain\_pow, medium\_went} \\

\bottomrule
\end{tabularx}
\end{table}

\textbf{Robustness analysis for clustering.}
To assess the robustness of the sub-clustering procedure, we repeated the full recursive clustering pipeline ten times for each conspiracy concept label and measured the stability of the resulting word neighborhoods across runs. Rather than comparing word sets directly using set-overlap measures like Jaccard similarity which penalizes heavily when cluster sizes differ across runs, we measure stability in the semantic vector space itself. Specifically, for each run, we compute the cosine similarity between the conspiracy concept label vector and each word in the final cluster produced for that term and average these scores to obtain a single stability score per run. A high cosine similarity indicates that final cluster words are semantically close to the target term across runs, regardless of which words were sampled at each iteration. 
We report the mean and standard deviation (SD) of this stability score across the ten runs for each term — a high mean and low standard deviation shows that the final cluster words are consistently semantically close to the target term regardless of sampling variation. For instance, \textit{sandy\_hook} achieved stability score of 0.95 (SD = 0.04) and \textit{chemtrails} 0.90 (SD  = 0.07),
suggesting that despite the randomness introduced by sampling at each iteration, the sub-clustering procedure reliably converges to the same semantic neighborhood around each conspiracy term.

\end{document}